\documentclass[11pt]{article}

\usepackage{emnlp2021}

\usepackage{times}
\usepackage{latexsym}

\usepackage[T1]{fontenc}

\usepackage[utf8]{inputenc}

\usepackage{microtype}

\usepackage{bm}
\usepackage{amsmath}
\usepackage{amssymb}
\usepackage{bbm}
\usepackage{graphicx}

\usepackage{algorithm}  
\usepackage{algorithmic} 
\usepackage{color}

\usepackage{multirow}
\usepackage{booktabs}

\title{A Role-Selected Sharing Network for Joint Machine-Human Chatting Handoff and Service Satisfaction Analysis}

\author{
    Jiawei Liu\textsuperscript{\rm 1},
    Kaisong Song\textsuperscript{\rm 2,3},
    Yangyang Kang\textsuperscript{\rm 3},
    Guoxiu He\textsuperscript{\rm 4},
    Zhuoren Jiang\textsuperscript{\rm 5},\\
    \bf{Changlong Sun\textsuperscript{\rm 3,5},
    Wei Lu\textsuperscript{\rm 1}\thanks{\quad Corresponding authors.}\ ,
    Xiaozhong Liu\textsuperscript{\rm 6}\footnotemark[1]}\\
    
    \textsuperscript{\rm 1}Wuhan University, Wuhan, China \quad
    \textsuperscript{\rm 2}Northeastern University, Shenyang, China\\
    \textsuperscript{\rm 3}Alibaba Group, Hangzhou, China \quad
    \textsuperscript{\rm 4}East China Normal University, Shanghai, China\\
    \textsuperscript{\rm 5}Zhejiang University, Hangzhou, China \quad
    \textsuperscript{\rm 6}Worcester Polytechnic Institute, Worcester, USA\\
    
    \texttt{\small{\{laujames2017,weilu\}@whu.edu.cn, \{kaisong.sks,yangyang.kangyy\}@alibaba-inc.com,}}\\
    \texttt{\small{gxhe@fem.ecnu.edu.cn, jiangzhuoren@zju.edu.cn,}}\\
    \texttt{\small{changlong.scl@taobao.com, xliu14@wpi.edu}}
}

\begin{document}
\maketitle
\begin{abstract}

Chatbot is increasingly thriving in different domains, however, because of unexpected discourse complexity and training data sparseness, its potential distrust hatches vital apprehension. Recently, Machine-Human Chatting Handoff (MHCH), predicting chatbot failure and enabling human-algorithm collaboration to enhance chatbot quality, has attracted increasing attention from industry and academia. In this study, we propose a novel model, Role-Selected Sharing Network (RSSN), which integrates both dialogue satisfaction estimation and handoff prediction in one multi-task learning framework. Unlike prior efforts in dialog mining, by utilizing local user satisfaction as a bridge, global satisfaction detector and handoff predictor can effectively exchange critical information. Specifically, we decouple the relation and interaction between the two tasks by the role information after the shared encoder. Extensive experiments on two public datasets demonstrate the effectiveness of our model.

\end{abstract}

\section{Introduction}

\begin{figure}[!t]
\centering
  \includegraphics[width=7.5cm]{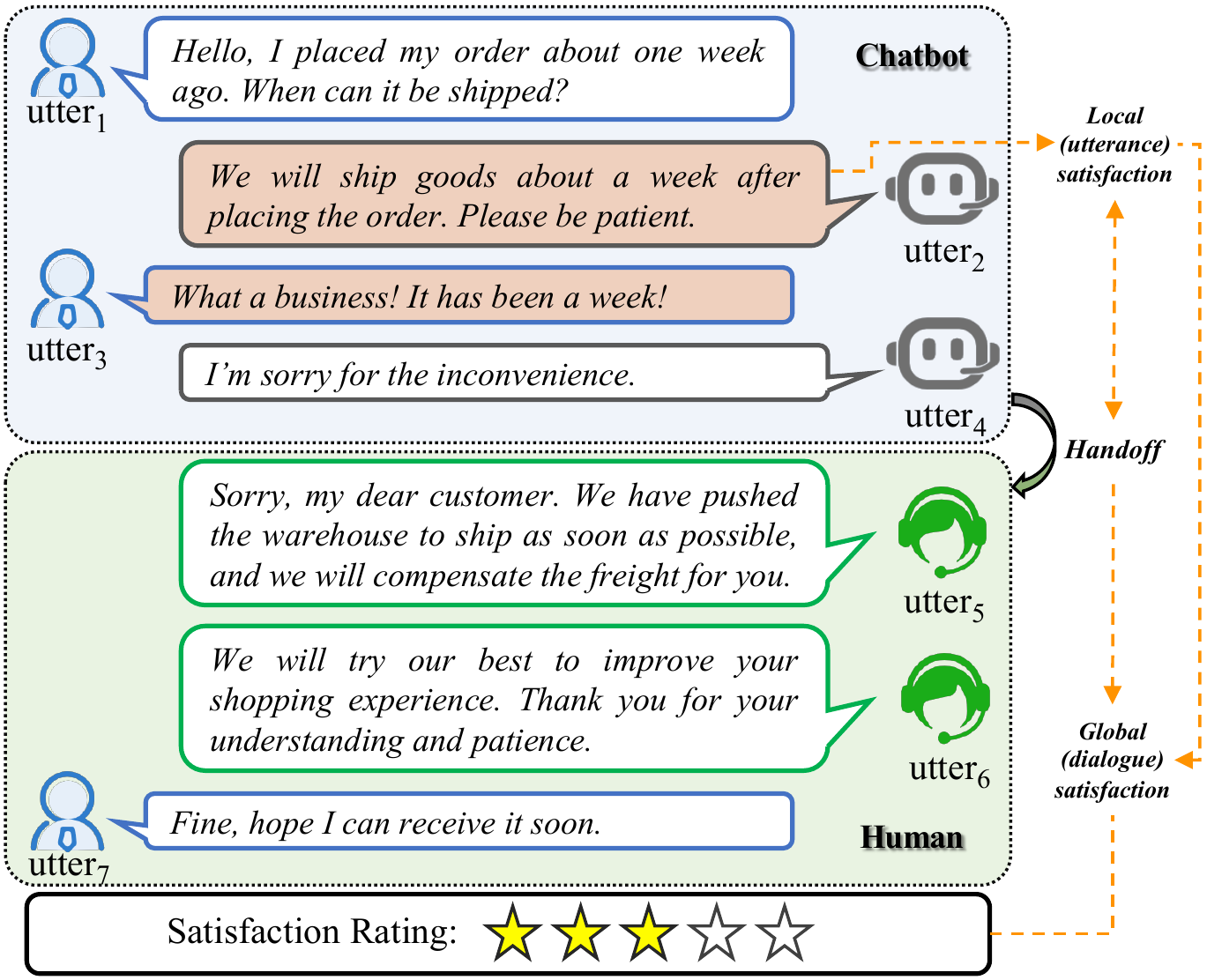} 
  \caption{A snippet of a moderately satisfied customer service dialogue. There is a satisfaction rating at the end of the conversation. The utterance with an orange background color denotes a transferable utterance.}
  \label{dialog_example} 
\end{figure}

Chatbot, as one of the recent palpable AI excitements, has been widely adopted to reduce the cost of customer service \citep{qiu2017alime,ram2018conversational,zhou2020design}. However, due to the complexity of human conversation, auto-chatbot can hardly meet all users' needs, while its potential failure perceives skepticism. AI-enabled customer service, for instance, may trigger unexpected business losses because of chatbot failures \citep{radziwill2017evaluating,rajendran2019learning}. Moreover, for chatbot adoption in sensitive areas, such as healthcare \citep{chung2019chatbot} and criminal justice \citep{wang2020masking}, any subtle statistical miscalculation may trigger serious health and legal consequences. To address this problem, recently, scholars proposed new dialog mining tasks to auto-assess dialogue satisfaction, a.k.a. Service Satisfaction Analysis (SSA) at dialogue-level \citep{song2019using}, and to predict potential chatbot failure via machine-human chatting handoff (MHCH) at utterance-level \citep{huang2018evorus,liu2021time}. In a MHCH context, algorithm can transfer an ongoing auto-dialogue to the human agent when the current utterance is confusing.

Figure \ref{dialog_example} depicts an exemplar dialogue of online customer service. In this dialogue, the chatbot gives an unsatisfied answer about shipping, thus causing the customer's complaint (local dissatisfaction $utter_2$ and $utter_3$). Ideally, chatbot should be able to detect the negative (local) emotion ($utter_3$) and tries to appease complaints, but this problem remains unresolved. If chatbot continues, the customer may cancel the deal and give a negative rating (dialogue global dissatisfaction). With MHCH (detects the risks of $utter_2$ and $utter_3$), the dialogue can be transferred to the human agent, who is better at handling, compensating, and comforting the customer and enhance customer satisfaction. This example illustrates the cross-impact between handoff and dialogue (local+global) satisfaction. Intuitively, MHCH and SSA tasks can be compatible and complementary given a dialogue discourse, i.e., the local satisfaction is related to the quality of the conversation \citep{bodigutla2019multi, bodigutla2020joint}, which can support the handoff judgment and ultimately affect the overall satisfaction. On the one hand, handoff labels of utterances are highly pertinent to local satisfaction, e.g., one can utilize single handoff information to enhance local satisfaction prediction, which ultimately contributes to the overall satisfaction estimation. On the other hand, the overall satisfaction is obtained by combining local satisfactions, which reflects the quality in terms of answer generation, language understanding, and emotion perception, and subsequently helps to facilitate handoff judgment.

In recent years, researchers \citep{bodigutla2019multi,bodigutla2019domain,ultes2019improving,bodigutla2020joint} explore joint evaluation of turn and dialogue level qualities in spoken dialogue systems.
In terms of general dialogue system, to improve the efficiency of dialogue management, \citet{qin2020dcr} propose a co-interactive relation layer to explicitly examine the cross-impact and model the interaction between sentiment classification and dialog act recognition, which are relevant tasks at the same level (utterance-level). However, MHCH (utterance-level) and SSA (dialogue-level) target satisfaction at different levels. More importantly, handoff labels of utterances are more comprehensive and pertinent to local satisfaction than sentiment polarities. Meanwhile, customer utterances have significant impacts on the overall satisfaction \citep{song2019using}, which motivates us that the role information can be critical for knowledge transfer of these two tasks.

To address the aforementioned issues, we propose an innovative Role-Selected Sharing Network (RSSN) for handoff prediction and dialogue satisfaction estimation, which utilizes role information to selectively characterize complex relations and interactions between two tasks. To the best of our knowledge, it is the pioneer investigation to leverage the multi-task learning approach for integrating MHCH and SSA. In practice, we first adopt a shared encoder to obtain the shared representations of utterances. Inspired by the co-attention mechanism \citep{xiong2016dynamic,qin2020dcr}, the shared representations are then fed into the role-selected sharing module, which consists of two directional interactions: \textit{MHCH to SSA} and \textit{SSA to MHCH}. This module is used to get the fusion of MHCH and SSA representations. We propose the role-selected sharing module based on the hypothesis that the role information can benefit the tasks' performances. The satisfaction distributions of utterances from different roles (agent and customer) are different, and the effects for the tasks are also different. Specifically, the satisfaction of agent is non-negative. The utterances from agent can enrich the context of customer's utterances and indirectly affect satisfaction polarity. Thus, directly employing local satisfaction of agent into the interaction with handoff may introduce noise. In the proposed role-selected sharing module, we adopt local satisfaction based on the role information: only the local satisfaction from customer can be adopted to interact with handoff information.
By this means, we can control knowledge transfer for both tasks and make our framework more explainable. The final integrated outputs are then fed to separate decoders for handoff and satisfaction predictions.

To summarize, our contributions are mainly as follows:
(1) We introduce a novel multi-task learning framework for combining machine-human chatting handoff and service satisfaction analysis. 
(2) We propose a Role-Selected Sharing Network for handoff prediction and satisfaction rating estimation, which can utilize different role information to control knowledge transfer for both tasks and enhance model performance and explainability.
(3) The experimental results demonstrate that our model outperforms a series of baselines that consists of the state-of-the-art (SOTA) models on each task and multi-task learning models for both tasks. To assist other scholars in reproducing the experiment outcomes, we release the codes and the annotated dataset\footnote{https://github.com/WeijiaLau/RSSN}.

\section{Related Work}
Due to the complexity of human conversation, current automatic chatbots are not mature enough and still fail to meet users' expectations \citep{brandtzaeg2018chatbots,jain2018evaluating,chaves2020should}. Besides exploring novel dialogue models, dialogue quality estimation, service satisfaction analysis, and human intervention are vital strategies to enhance chatbot performance.

\textbf{Dialogue Quality and Service Satisfaction Analysis}.
Interaction Quality (IQ) \citep{schmitt2012parameterized} and Response Quality (RQ) \citep{bodigutla2019domain} are dialogue quality evaluation metrics for spoken dialogue systems. Automated models to estimate IQ \citep{ultes2014application,el2014ordinal} and RQ \citep{bodigutla2019multi,bodigutla2019domain,bodigutla2020joint} utilize various features derived from the dialogue content and output from spoken language understanding components. For chat-oriented dialogue system, \citet{higashinaka2015towards,higashinaka2015fatal} introduce Dialogue Breakdown Detection task to detect a system's inappropriate utterances that lead to dialogue breakdowns. To efficiently analyze dialogue satisfaction, \citet{song2019using} introduce the task of service satisfaction analysis (SSA) based on multi-turn customer service dialogues. The proposed CAMIL model can predict the sentiment of all the customer utterances and aggregate those sentiments into overall service satisfaction polarity. Nevertheless, the sentiment of customer utterance is only one of the factors that influence service satisfaction.

\textbf{Machine Human Chatting Handoff}.
Another perspective of further enhancing the chatbot's performance is to combine chatbots with human agent. Recently, there are several works about human-machine cooperation for chatbots. \citet{huang2018evorus} propose the crowd-powered conversational assist architecture, namely Evorus, which integrates crowds with multiple chatbots and a voting system. \citet{rajendran2019learning} utilize reinforce learning framework to transfer conversations to human agents once encountered new user behaviors. Different from them, \citet{liu2021time} mainly focus on detecting transferable utterances which are one of the keys to improve user satisfaction. They propose a DAMI network that utilizes difficulty-assisted encoding and matching inference mechanisms to predict the transferable utterance.

\textbf{Multi-task learning in dialogue system}.
For satisfaction estimation, \citet{bodigutla2020joint} propose to jointly predict turn-level RQ labels and dialogue-level ratings. They utilize features from spoken dialogue system and BiLSTM \citep{hochreiter1997long} based model to automatically weight each turn's contribution towards the rating. \citet{ma2018detect} propose a joint framework that unifies two highly pertinent tasks. Both tasks are trained jointly using weight sharing to extract the common and task-invariant features while each task can still learn its task-specific features. To learn the correlation between two tasks, \citet{qin2020dcr} propose a DCR-Net. It adopts a stacked co-interactive relation layer to incorporate mutual knowledge explicitly. This model ignores the contextual information and isolated two types of information when performing interaction.

\section{Methodology}
\begin{figure*}[!ht]
\centering
\includegraphics[width=14.9cm]{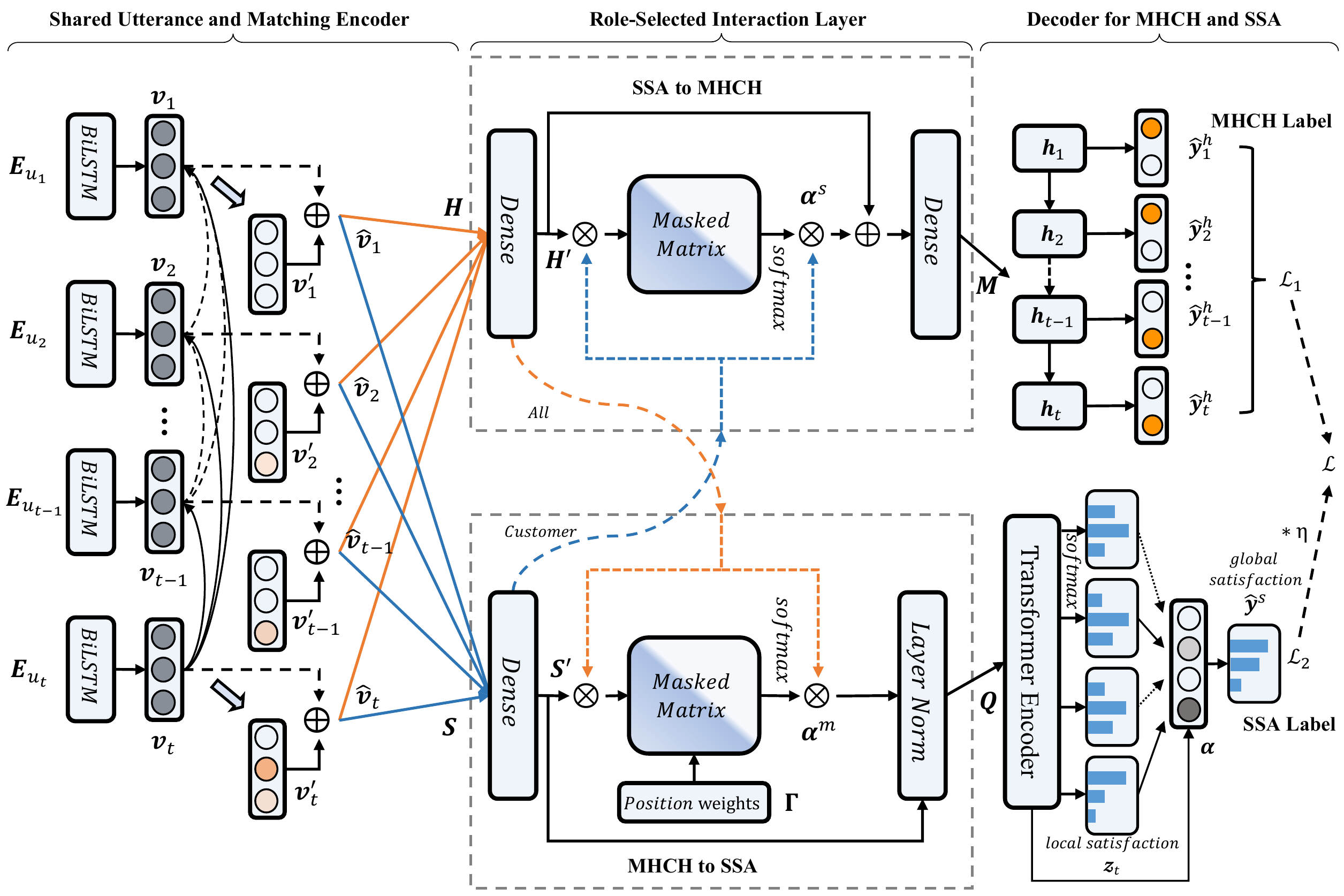} 
\caption{The architecture of our Role-Selected Sharing Network (RSSN) Network.}
\label{fig:rssn}
\end{figure*}
Figure \ref{fig:rssn} shows the overall architecture of RSSN, which consists of three parts: Shared Utterance and Matching Encoder, Role-Selected Interaction Layer, and Decoder for MHCH and SSA. In this section, we will describe them in detail.

Given a dialogue $D=[u_1,..., u_L]$, it consists of a sequence of $L$ utterances with corresponding handoff labels $[y^{h}_1,..., y^{h}_L]$, where $t=\{1\leq t \leq L | t\in \mathbb{N}\}$, ${y^{h}_t \in \Psi}$ and $\Psi=\{$\textit{normal, transferable}$\}$. \textit{Transferable} indicates the dialogue should be transferred to the human agent, whereas \textit{normal} indicates there is no need to transfer. The satisfaction polarity of dialogue $D$ is noted as $y^s$, where $y^s \in \Omega$ and $\Omega = \{$\textit{well satisfied, met, unsatisfied}$\}$.
Note that we perform the multi-task learning with the supervision of handoff labels and dialogue's satisfaction only. The local satisfaction distributions of utterances are only the latent estimation, which helps to predict the dialogue's satisfaction.

\subsection{Shared Utterance and Matching Encoder}
The shared encoder consists of a bidirectional LSTM (BiLSTM) to learn the utterance representation and a masked matching layer to capture the contextual matching information.

Suppose ${{u}_{t}=[ w_1, ..., w_{\left | u_t\right |}]}$ represents a sequence of words in the $t$-th utterance. These words are mapped into corresponding word embeddings ${\boldsymbol{E}_{u_t}\in \mathbb{R}^{n\times \left | u_t\right |}}$, where $n$ is the word embedding dimension. By adopting semantic composition models with word embeddings, we can learn the utterance representation. In this work, we adopt a BiLSTM model and concatenate hidden states of forward and backward LSTM to learn the context-sensitive utterance representation $\boldsymbol{v}_{t}\in \mathbb{R}^{2k}$, where $k$ is the number of hidden units of LSTM cell. Formally, we have $\boldsymbol{v}_{t} = \text{BiLSTM}(\boldsymbol{E}_{u_t})$.

In a dialogue, preceding utterances for each utterance provide helpful context information to estimate local satisfaction. Thus, within a dialogue, there is a high probability of inter-dependency with respect to their context clues. To encapsulate the contextual matching and information flow in the dialogue, we feed the utterance representation into a unidirectional matching mechanism:
\begin{align}
    {\boldsymbol{v}^{\prime}_{t} = \boldsymbol{v}^{\top}_{t} [\boldsymbol{v}_{1},\boldsymbol{v}_{2},...,\boldsymbol{v}_{t-1}]}
\end{align}

After masking out the future information of the present utterance, the matching features of dialogue $D$ is a lower triangular matrix with the diagonal values removed. Then we concatenate the matching features with utterance representation to get $\hat{\boldsymbol{v}}_{t} = [\boldsymbol{v}^{\prime}_{t}; \boldsymbol{v}_t]$.
Finally, we obtain the initial shared utterances representations of MHCH $\boldsymbol{H} = [\hat{\boldsymbol{v}}_1,...,\hat{\boldsymbol{v}}_L]$ and SSA $\boldsymbol{S}=[\hat{\boldsymbol{v}}_1,...,\hat{\boldsymbol{v}}_L]$.

\subsection{Role-Selected Interaction Layer}
In customer service dialogue, the roles of different participants would exhibit different characteristics \citep{song2019using}. Besides, we conjecture that MHCH and SSA have different impacts on each other. These two tasks indirectly establish a connection through various factors such as dialogue quality, satisfaction, and sentiment. At the same time, role information also plays an important role in both tasks. On the one hand, the utterances from agent can enrich the context of customer utterances and indirectly affect satisfaction polarity. In contrast, customer utterances tend to have a more direct impact on the dominating satisfaction polarity. On the other hand, the utterances of any participants can trigger machine-human chatting handoff. Thus, we propose the Role-Selected Interaction Layer, which contains two interaction directions: \textit{SSA to MHCH} and \textit{MHCH to SSA}, to model the relations and interactions between the two tasks separately.

We first apply two Dense layers over the handoff information and satisfaction information respectively to make them more task-specific, which can be noted as $\boldsymbol{H}^{\prime} = \text{Dense}(\boldsymbol{H})$ and $\boldsymbol{S}^{\prime} = \text{Dense}(\boldsymbol{S})$, where $\boldsymbol{H}^{\prime} \in \mathbb{R}^{L \times d}$ and $\boldsymbol{S}^{\prime} \in \mathbb{R}^{L \times d}$. Note that $d$ is the number of hidden units of the Dense layer.

\textbf{SSA to MHCH}. Co-attention is an effective and widely used method to capture the mutual knowledge among the correlated tasks \citep{xiong2016dynamic, qin2020dcr}. Inspired by the basic co-attention mechanism, we design the interaction mechanism separately according to the characteristics of tasks. In this way, task-relevant knowledge can be transferred mutually between two tasks. Specifically, the \textit{SSA to MHCH} module produces comprehensive handoff representations incorporating the local satisfaction information. Since the agent utterances indirectly affect satisfaction polarity, directly employing local satisfaction of agent into the interaction with handoff may introduce noise. As a consequence, we only adopt the local satisfaction information of customer to interact with handoff information. The process can be defined as follows:
\begin{gather}
    {\boldsymbol{\alpha}}^{s} = \text{softmax}({\text{Mask}_c}(\boldsymbol{H}^{\prime}(\boldsymbol{S}^{\prime})^{\top}))\\
    \boldsymbol{M} = \text{Dense}( [\boldsymbol{{\alpha}}^{s}\boldsymbol{S}^{\prime};\boldsymbol{H}^{\prime}])
\end{gather}

where $\boldsymbol{M} \in \mathbb{R}^{L \times d}$ and $\text{Mask}_{c}$ denotes that we mask out (setting to $-\infty$) all values of the future information and agent utterances.

\begin{figure}[!t]
\centering
  \includegraphics[width=7.8cm]{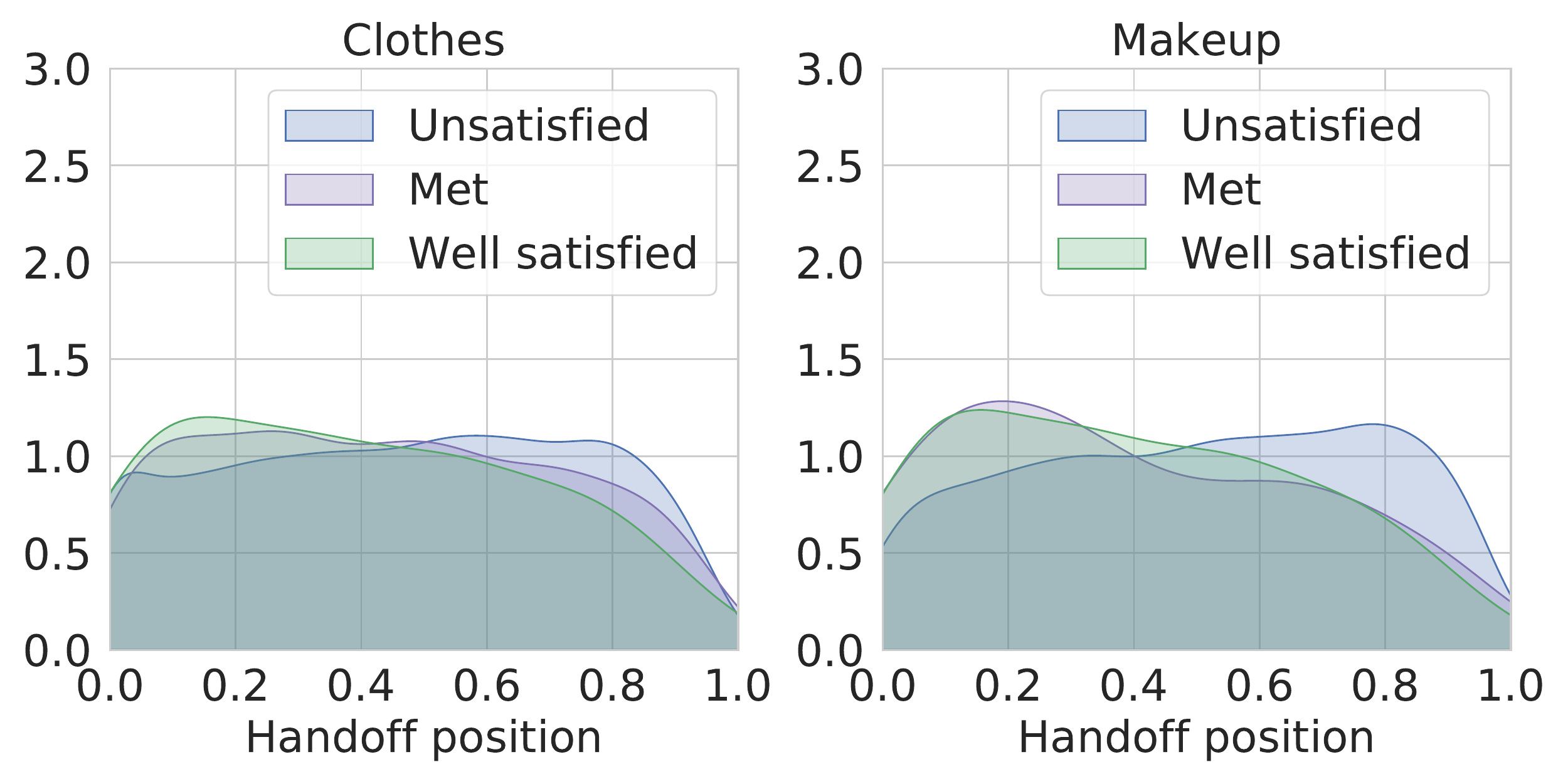}
  \caption{The relative handoff position distributions in three different service satisfaction ratings.}
  \label{handoff_distri} 
\end{figure}

\textbf{MHCH to SSA}. As shown in Figure \ref{handoff_distri}, we observe that the dialogue satisfaction rating is related to the handoff position. Intuitively, a handoff can be triggered by the local unsatisfied attitude of the customer, and the later handoff means users are unsatisfied before the end of the conversation. Prior study, \citet{song2019using}, also found that user satisfaction at the dialogue level is usually determined by the attitudes of the last few utterances. We can derive that handoff at the later period of the conversation may result in a lower satisfaction rating. Thus, we adjust the interactive attention by positional weights, which can be computed as below:

\begin{align}
    {\boldsymbol{\beta}_t} = \text{softmax}([\frac{1}{L},...,\frac{t}{L},...,1]\odot \boldsymbol{I}_{p}(u_t))
\end{align}
where $\odot$ is element-wise product and $I_p(\cdot)$ denotes a zero masking identity matrix to mask out future information. Finally, the positional weights $\boldsymbol{\Gamma } = [{\boldsymbol{\beta}_1};...;{\boldsymbol{\beta}_{L}}]$, where $\boldsymbol{\Gamma } \in \mathbb{R}^{L\times L}$. The mechanism gives more weight to the later handoff information.
We apply the positional weights to the interaction:
\begin{gather}
    \boldsymbol{\alpha}^{m} = \text{softmax}(\text{Mask}( \boldsymbol{S}^{\prime} \cdot (\boldsymbol{H}^{\prime} )^{\top} \cdot \mathbf{\Gamma }))\\
    \boldsymbol{Q} = \text{LayerNorm}(\boldsymbol{\alpha}^{m} \cdot \boldsymbol{H}^{\prime} + \boldsymbol{S}^{\prime})
\end{gather}
where $\text{Mask}$ denotes that we mask out the future information (setting to $-\infty$), and $\text{LayerNorm}$ denotes the layer normalization \citep{ba2016layer}.

\subsection{Decoder for MHCH and SSA}
After the role-selected interaction layer, we can get the outputs $\boldsymbol{M}=[\boldsymbol{m}_1,...,\boldsymbol{m}_L]$ and $\boldsymbol{Q}=[\boldsymbol{q}_1,...,\boldsymbol{q}_L]$. Then we adopt separate decoders to predict handoff and satisfaction rating.

In terms of machine-human chatting handoff, the tendency of handoff also depends on the dialogue context. Thus, we feed the outputs of the interaction layer into an LSTM to connect the sequential information flow in the dialogue:
\begin{align}
    {\boldsymbol{h}_t = \text{LSTM}(\boldsymbol{m}_{t}, \boldsymbol{h}_{t-1})}
\end{align}
where $\boldsymbol{h}_t \in \mathbb{R}^{k}$ is the hidden state for $u_t$. Since there are no dependencies among labels, we simply use a softmax classifier for handoff prediction:
\begin{align}
    \hat{\boldsymbol{y}}^{h}_{t} = \text{softmax}(\boldsymbol{\textit{W}}_{\hbar} \boldsymbol{h}_t + \boldsymbol{b}_{\hbar}) 
\end{align}
where $\boldsymbol{\textit{W}}_{\hbar} \in \mathbb{R}^{|\Psi| \times k}$ and $\boldsymbol{b}_{\hbar} \in \mathbb{R}^{|\Psi|}$. $\hat{\boldsymbol{y}}^{h}_{t} \in \mathbb{R}^{|\Psi|}$ is the predicted handoff probability distribution of $u_t$.

For service satisfaction analysis, we first apply a transformer block \citep{vaswani2017attention} to model the long-range context of the dialogue further. Formally, we have $\hat{\boldsymbol{Q}} = \text{Transformer}(\boldsymbol{Q})$, where $\hat{\boldsymbol{Q}}=\{\hat{\boldsymbol{q}}_1,...,\hat{\boldsymbol{q}}_L | \hat{\boldsymbol{q}}_t \in \mathbb{R}^{k}\}$.

Then we utilize a softmax function for estimating local satisfaction distribution $\boldsymbol{z}_t \in \mathbb{R}^{|\Omega|}$ of $u_t$:
\begin{align}
    \boldsymbol{z}_t = \text{softmax}(\boldsymbol{\textit{W}}_{\xi} \hat{\boldsymbol{q}}_t + \boldsymbol{b}_{\xi}) 
\end{align}
where $\boldsymbol{\textit{W}}_{\xi} \in \mathbb{R}^{{|\Omega|} \times k}$ and $\boldsymbol{b}_{\xi} \in \mathbb{R}^{|\Omega|}$.

Since only a fraction of customer utterances can contribute to the final satisfaction rating, we introduce an attention strategy that enables our model to attend to customer utterances of different importance when merging the local satisfaction distribution.
Formally, we measure the importance of each customer utterances as below:
\begin{align}
    \boldsymbol{\alpha} = \text{softmax}(\text{Mask}^{\prime}_{c}( \boldsymbol{g}^\top \text{tanh}(\boldsymbol{\textit{W}}_{\mu} {\hat{\boldsymbol{Q}}}^\top + \boldsymbol{b}_{\mu})))
\end{align}
where $\boldsymbol{\alpha} \in \mathbb{R}^L$. $\boldsymbol{\textit{W}}_{\mu} \in \mathbb{R}^{z \times k}$, $\boldsymbol{b}_{\mu} \in \mathbb{R}^{z}$, and $\boldsymbol{g} \in \mathbb{R}^{z}$ are trainable parameters. $z$ is the number of attention units. $\text{Mask}^{\prime}_{c}$ denotes the masking function used to reserve customer utterances. $\boldsymbol{g}$ can be perceived as a high-level representation of a fixed query "Which is the critical utterance?".

Finally, we obtain the overall satisfaction distribution $\hat{\boldsymbol{y}}^{s} \in \mathbb{R}^{|\Omega|}$  as the weighted sum of local customer satisfaction distribution:
\begin{align}
    \hat{\boldsymbol{y}}^{s} = \sum^{L}_{t=1} {\alpha}_{t} \boldsymbol{z}_t
\end{align}
where $\alpha_t$ is the $t$-th weight of utterance $u_t$ in $\boldsymbol{\alpha}$.

\subsection{Joint Training}
The objective function of MHCH is formulated as:
\begin{align}
    \mathcal{L}_1 = - \frac{1}{L}\sum_{t=1}^{L} \sum_{i=1}^{|\Psi|}{y}^{h}_{i,t} \log(\hat{{y}}^{h}_{i,t})
\end{align}
The objective function of SSA is formulated as:
\begin{align}
    \mathcal{L}_2 = - \sum_{i=1}^{|\Omega|} {y}^{s}_{i} \log(\hat{{y}}^{s}_{i})
\end{align}

Finally, we minimize the joint cross-entropy loss $\mathcal{L}$, which is obtained as follow:
\begin{align}
    \mathcal{L}(\Theta) = \mathcal{L}_1 + \eta * \mathcal{L}_2 +\delta \left \| \Theta \right \|_2^2
\end{align}
where $\eta \in \mathbb{R}^{+}$ denotes the trade-off parameter, $\delta$ denotes the $L_2$ regularization weight, and $\Theta$ denotes all the trainable parameters of model. We use backpropagation to compute the gradients of the parameters, and update them with Adam \citep{kingma2014adam} optimizer.

\section{Experiments and Results}

\subsection{Dataset and Experimental Settings}

\begin{table}[!t]
\small
  \centering
    \begin{tabular}{lcc}
    \toprule
    \multicolumn{1}{c}{\textbf{Statistics items}} & \textbf{Clothes} & \textbf{Makeup} \\
    \midrule
    \midrule
    \# Dialogues & 10,000 & 3,540\\
    \# US (unsatisfied) & 2,302 & 1,180 \\
    \# MT (met) & 6,399 & 1,180 \\
    \# WS (well satisfied) & 1,299 & 1,180 \\
    \# Transferable Utterances & 16,921 & 7,668 \\
    \# Normal Utterances & 237,891 & 86,778 \\
    Avg \# Utterances & 25.48 & 26.67 \\
    Avg \# Tokens & 7.64  & 7.87 \\
    \midrule
    Kappa & 0.85 & 0.88 \\
    \bottomrule
    \end{tabular}%
  \caption{Statistics of the datasets.}
  \label{tab:stat}%
\end{table}%

Our experiments are conducted based on two publicly available Chinese customer service dialogue datasets, namely \textbf{Clothes} and \textbf{Makeup}\footnote{https://github.com/songkaisong/ssa}, collected by \citet{song2019using} from Taobao\footnote{https://www.taobao.com}. Both datasets have service satisfaction ratings from customer feedbacks and annotated sentiment labels of utterances. Note that the sentiment labels do not participate in our training process and are only used for test. Meanwhile, we also annotate the \textit{transferable/normal} labels for both datasets according to the existing specifications \citep{liu2021time}. Two annotators with professional linguistics knowledge participated in the annotation task.

A summary of statistics, including Kappa value \citep{snow2008cheap} for both datasets are given in Table \ref{tab:stat}. Clothes is a corpus with 10K dialogues in the \textit{Clothes} domain, which has an imbalanced satisfaction distribution at dialogue level. Makeup is a corpus with 3,540 dialogues in the \textit{Makeup} domain, which has a balanced satisfaction distribution dialogue level.
Note that we do not adopt the original word segmentation.
Figure \ref{handoff_distri} shows the relative handoff position distributions in different satisfaction ratings, where we take explicit request, negative emotion, and unsatisfactory answer handoffs into consideration. It indicates that handoff at the later phase of the conversation is more likely to get a lower service satisfaction rating.

Except BERT-based model, all texts are tokenized by a popular Chinese word segmentation utility called jieba\footnote{https://pypi.org/project/jieba}. The datasets are partitioned for training, validation, and test with an 80/10/10 split. For the BERT-based methods, we fine-tune the pre-trained model. For the other methods, we apply the pre-trained word vectors initially trained on Clothes and Makeup corpora by using CBOW \citep{mikolov2013efficient}. The dimension of word embedding is set as 200. Other trainable model parameters are initialized by sampling values from the Glorot uniform initializer \citep{glorot2010understanding}.
The sizes of hidden state $k$, Dense units $d$, attention units $z$, and batch size are selected from $\{32, 64, 128, 256, 512\}$. The dropout \citep{srivastava2014dropout} rate and the loss weight $\eta$ are selected from $(0,1)$ by grid search. Finally, we train the models with an initial learning rate of $1.5\times10^{-3}$ and $2\times10^{-5}$ for regular baselines and BERT-based models. All the methods run on a server configured with a Tesla V100, 32 CPU, and 32G memory.

\begin{table*}[htbp]
\renewcommand\tabcolsep{2.5pt} 
  \centering
  
  \resizebox{1\textwidth}{!}{
    \begin{tabular}{l|ccccc|ccccc|ccccc|ccccc}
    \toprule
    \multirow{3}[6]{*}{\textbf{Models}} & \multicolumn{10}{c|}{\textbf{Clothes}}                                        & \multicolumn{10}{c}{\textbf{Makeup}} \\
\cmidrule{2-21}          & \multicolumn{5}{c|}{\textbf{MHCH}}    & \multicolumn{5}{c|}{\textbf{SSA}}     & \multicolumn{5}{c|}{\textbf{MHCH}}    & \multicolumn{5}{c}{\textbf{SSA}} \\
\cmidrule{2-21}          & F1    & Mac. F1 & GT-I  & GT-II & GT-III & WS F1 & MT F1 & US F1 & Mac. F1 & Acc.  & F1    & Mac. F1 & GT-I  & GT-II & GT-III & WS F1 & MT F1 & US F1 & Mac. F1 & Acc. \\
    \midrule
    \midrule
    HAN   & 59.8  & 78.7  & 71.7  & 73.1  & 74.0  & 51.5  & 81.7  & 70.4  & 67.9  & 75.5  & 54.3  & 75.4  & 68.5  & 70.1  & 71.3  & 68.4  & 71.3  & 84.8  & 74.8  & 74.8 \\
    BERT+LSTM & 60.4  & 78.9  & 73.4  & 74.9  & 75.9  & 42.2  & 84.2  & 72.9  & 66.4  & 77.6  & 59.1  & 78.0  & 72.0  & 73.0  & 73.7  & 66.7  & 72.9  & 87.2  & 75.6  & 76.0 \\
    \midrule
    HEC   & 59.8  & 78.7  & 71.2  & 72.3  & 73.0  & -     & -     & -     & -     & -     & 57.1  & 76.8  & 68.0  & 69.5  & 70.5  & -     & -     & -     & -     & - \\
    DialogueRNN & 60.8  & 79.2  & 73.1  & 74.6  & 75.6  & -     & -     & -     & -     & -     & 58.3  & 77.4  & 68.8  & 70.5  & 71.6  & -     & -     & -     & -     & - \\
    CASA  & 62.0  & 79.8  & 73.6  & 75.0  & 75.9  & -     & -     & -     & -     & -     & 58.4  & 77.5  & 70.6  & 72.7  & 73.9  & -     & -     & -     & -     & - \\
    LSTMLCA & 62.6  & 80.1  & 72.4  & 73.9  & 74.8  & -     & -     & -     & -     & -     & 57.4  & 77.0  & 70.2  & 71.7  & 72.6  & -     & -     & -     & -     & - \\
    CESTa & 60.6  & 79.1  & 73.4  & 74.8  & 75.6  & -     & -     & -     & -     & -     & 59.3  & 78.0  & 69.6  & 71.2  & 72.2  & -     & -     & -     & -     & - \\
    DAMI  & \underline{66.7}  & \underline{82.2}  & 74.2  & 75.9  & \underline{77.1}  & -     & -     & -     & -     & -     & \underline{61.1}  & \underline{79.0}  & \underline{73.3}  & \underline{74.4}  & \underline{75.2}  & -     & -     & -     & -     & - \\
    \midrule
    MILNET & -     & -     & -     & -     & -     & 38.2  & 82.3  & 70.8  & 63.8  & 75.3  & -     & -     & -     & -     & -     & 72.0  & 68.9  & 84.9  & 75.3  & 75.1 \\
    HMN   & -     & -     & -     & -     & -     & 44.1  & 83.3  & 69.6  & 65.7  & 76.3  & -     & -     & -     & -     & -     & 73.5  & 73.1  & 83.4  & 76.6  & 76.8 \\
    CAMIL & -     & -     & -     & -     & -     & \underline{55.4}  & \underline{84.4}  & 71.5  & \underline{70.4}  & \underline{78.3}  & -     & -     & -     & -     & -     & 73.8  & \underline{74.5}  & 87.4  & \underline{78.6}  & \underline{78.5} \\
    \midrule
    MT-ES   &   61.7 &  79.7 &  74.6 &  75.9 &  76.8 &  47.7 &  82.4 &  74.1 &  68.1 &  76.4 &  57.1 &  76.9 &  69.9 &  71.7 &  72.8 &  72.0 &  68.7 &  84.3 &  75.0 &  75.1 \\
    JointBiLSTM &   62.0 &  79.9 &  \underline{75.0} &  \underline{76.1} &  76.9 &  26.7 &  82.1 &  69.4 &  59.4 &  74.5 &  59.3 &   78.0 &    70.1 &  72.0 &  73.1 &  74.5 &  72.2 &  83.7 &  76.8 &  76.8 \\
    DCR-Net & 62.1  & 79.9  & 71.4  & 72.8  & 73.7  & 49.8  & 82.7  & \textbf{\underline{76.6}}  & 69.7  & 77.3  & 58.8  & 77.7  & 70.0  & 72.1  & 73.4  & \underline{74.8}  & 69.1  & \underline{88.6}  & 77.5  & 77.7 \\
    \midrule
    \textbf{RSSN(ours)} & \textbf{69.2}$^{*}$ & \textbf{83.6}$^{*}$ & \textbf{78.4}$^{*}$ & \textbf{79.5}$^{*}$ & \textbf{80.3}$^{*}$ & \textbf{56.0}  & \textbf{85.1} & 74.0  & \textbf{71.7}$^{*}$ & \textbf{79.5}$^{*}$ & \textbf{65.9}$^{*}$ & \textbf{81.5}$^{*}$ & \textbf{75.1}$^{*}$ & \textbf{76.6}$^{*}$ & \textbf{77.6}$^{*}$ & \textbf{77.4}$^{*}$ & \textbf{76.1}$^{*}$ & \textbf{88.9}  & \textbf{80.8}$^{*}$ & \textbf{80.8}$^{*}$ \\
    \bottomrule
    \end{tabular}%
    }
    \caption{Experimental results of performance (\%) comparison with base models on Makeup and Clothes test datasets. \underline{\textit{Underline}} shows the best performance for baselines. - means not applicable. \textbf{\textit{Bold}} shows the best performance. ${}^{*}$ indicates statistical significance at $p<$ 0.05 level compared to the best performance of baselines.}
  \label{tab:compare}%
\end{table*}%

\subsection{Baselines}
We compare our model with 14 strong dialogue classification baseline models, which come from MHCH, SSA, and other similar tasks.

\textit{Generic Baselines:} \textbf{HAN} \citep{yang2016hierarchical} and \textbf{BERT}\citep{devlin2019bert}\textbf{+LSTM}.
We adopt outputs and the last hidden of RNN to predict handoff labels and the satisfaction rating, respectively.

\textit{Baselines for the MHCH task:} \textbf{HEC} \citep{kumar2017dialogue}, \textbf{DialogueRNN} \citep{majumder2019dialoguernn}, \textbf{CASA} \citep{raheja2019dialogue}, \textbf{LSTMLCA} \citep{dai2020local}, \textbf{CESTa} \citep{wang2020contextualized}, and \textbf{DAMI} \citep{liu2021time}.

\textit{Baselines for the SSA task:} \textbf{MILNET} \citep{angelidis2018multiple}, \textbf{HMN} \citep{shen2018sentiment}, and \textbf{CAMIL} \citep{song2019using}.

\textit{Multi-task baselines:} \textbf{MT-ES} \citep{ma2018detect}, \textbf{JointBiLSTM} \citep{bodigutla2020joint}, and \textbf{DCR-Net} \citep{qin2020dcr}. Specifically, We modify \textbf{DCR-Net} for our tasks by keeping the core self-attention and co-interactive relation layer.

For \textit{DAMI}, we adopt the open-sourced code\footnote{https://github.com/WeijiaLau/MHCH-DAMI} to get the results. For \textit{DialogueRNN}, we adapt the open-sourced code\footnote{https://github.com/senticnet/conv-emotion} to MHCH by keeping the core component unchanged. For \textit{HAN, MILNET, HMN}, and \textit{CAMIL} of SSA, we adopt the reported results from \citet{song2019using}. We re-implement the other models. For \textit{BERT+LSTM}, we adopt Chinese BERT-base model\footnote{https://github.com/google-research/bert}.

\subsection{Comparative Study}
Following \citet{song2019using}, we adopt \textit{Macro F1} (\textit{Mac. F1}) and \textit{Accuracy} (\textit{Acc.}) for evaluating the SSA task. For evaluating the MHCH task, we adopt \textit{F1}, \textit{Macro F1} (\textit{Mac. F1}), and \textit{Golden Transfer within Tolerance (GT-T)} \citep{liu2021time}. \textit{GT-T} considers the tolerance property of the MHCH task by the tolerance range $T$, which allows a ``biased'' prediction within it. The adjustment coefficient $\lambda$ of \textit{GT-T} penalizes early or delayed handoff. Likewise, we set $\lambda$ as 0, and set $T$ to range from 1 to 3 corresponding to \textit{GT-I}, \textit{GT-II}, and \textit{GT-III}. The results of comparisons are shown in Table \ref{tab:compare}.

We can observe that: (1) The proposed method outperforms all state-of-the-art models specific to one task in terms of all metrics on two datasets. This indicates that our proposed model can effectively capture useful information in both tasks by utilizing role and positional information to explicitly control the interaction between the two tasks. Hence, the performance of the two tasks can be boosted mutually.
(2) By integrating MHCH with SSA, the multi-task learning model can obtain further improvements. Specifically, we find that the MHCH task has a positive influence on detecting the unsatisfied dialogue. Overall, DCR-Net and our model perform better than standalone models on \textit{US F1} of satisfaction prediction. Intuitively, it is mainly because the interaction with handoff can more comprehensively reflect the local dissatisfaction dialogues than solely sentiment polarity analysis, which helps the joint model better identify dissatisfied dialogues for the SSA task.

\renewcommand\tabcolsep{2pt}
\begin{table}[t]
  \centering
  \resizebox{0.48\textwidth}{!}{
    \begin{tabular}{l|ccc|cc|ccc|cc}
    \toprule
    \multirow{3}[6]{*}{\textbf{Models}} & \multicolumn{5}{c|}{\textbf{Clothes}} & \multicolumn{5}{c}{\textbf{Makeup}} \\
\cmidrule{2-11}          & \multicolumn{3}{c|}{\textbf{MHCH}} & \multicolumn{2}{c|}{\textbf{SSA}} & \multicolumn{3}{c|}{\textbf{MHCH}} & \multicolumn{2}{c}{\textbf{SSA}} \\
\cmidrule{2-11}          & F1    & Mac. F1 & GT-I  & Mac. F1 & Acc.  & F1    & Mac. F1 & GT-I  & Mac. F1 & Acc. \\
    \midrule
    \midrule
    Average & 65.3  & 81.6  & 74.6  & 63.8  & 73.7  & 64.1  & 80.4  & 73.2  & 73.6  & 73.7 \\
    Voting & 61.7  & 79.7  & 71.7  & 27.5  & 61.2  & 62.0  & 79.4  & 68.3  & 34.6  & 42.1 \\
    Last  & 66.4  & 82.1  & 74.6  & 67.0  & 75.6  & 62.8  & 79.8  & 71.5  & 76.6  & 76.8 \\
    w/o Interact & 65.6  & 81.6  & 70.6  & 65.5  & 72.3  & 62.0  & 79.5  & 71.4  & 74.8  & 74.6 \\
    w/o Select & 64.4  & 81.0  & 73.6  & 67.0  & 74.3  & 61.7  & 79.2  & 71.7  & 72.9  & 72.9 \\
    w/o Position & 66.1  & 81.9  & 73.2  & 68.4  & 76.0  & 64.7  & 80.8  & 73.5  & 76.8  & 76.8 \\
    \midrule
    Full Model & 69.2  & 83.6  & 78.4  & 71.7  & 79.5  & 65.9  & 81.5  & 75.1  & 80.8  & 80.8 \\
    \bottomrule
    \end{tabular}%
    }
    \caption{Ablation study performance (\%) on Clothes and Makeup test datasets. w/o denotes "without".}
  \label{tab:ablation}%
\end{table}%

\begin{figure*}[t]
\centering
  \includegraphics[width=16cm]{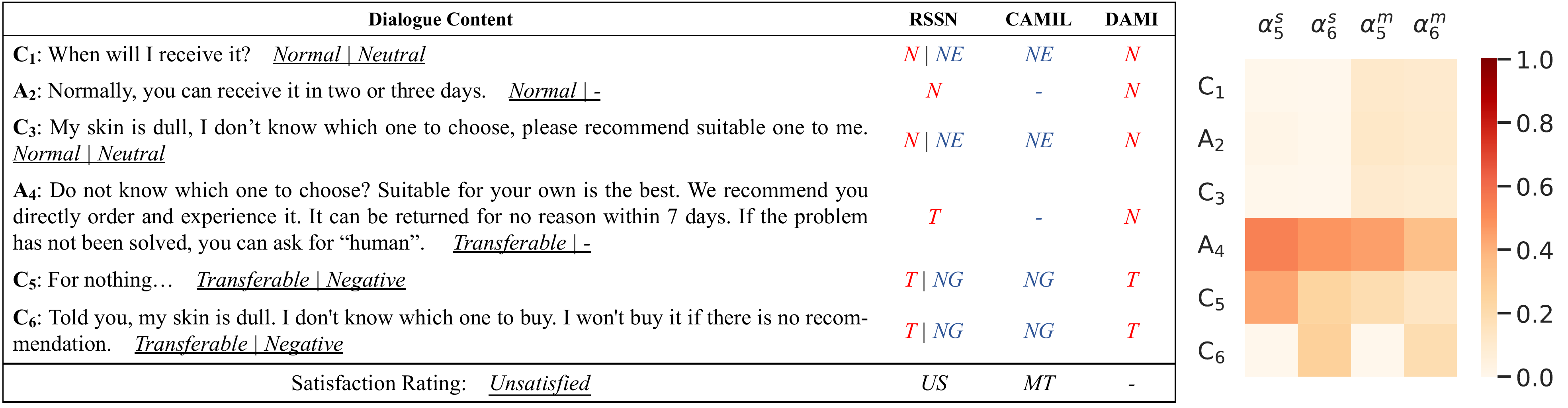} 
  \caption{An example dialogue with predictions and attention distribution. \textbf{C$_i$}/\textbf{A$_i$} denotes Customer/Chatbot utterance, followed by true \underline{\textit{labels}}. The sentiment labels of Customer utterances are also given along with the handoff labels. The other columns are the predictions of our model, CAMIL and DAMI, respectively. The satisfaction ratings of ground truth and predictions are in the last row of the table. N/T denotes \textit{Normal}/\textit{Transferable}.}
  \label{case} 
\end{figure*}

\subsection{Ablation}
We perform several ablation tests in our model on two datasets and the results are recorded in Table \ref{tab:ablation}. The results demonstrate the effectiveness of different components of our model.

\textbf{w/o Interact:} We modify the full version of our model by only sharing parameters of the Utterance and Matching Encoder. The performance degradation demonstrates the effectiveness of modeling the relations between two tasks with interaction.
\textbf{w/o Select:} We remove the Role-Select mechanism to ignore the role information during the interaction process. The performance degradation indicates that straightforward interaction may bring noisy information for both tasks.
\textbf{w/o Position}: We remove the positional weights in the MHCH to SSA sub-module. It performs well but worse than \textbf{Full Model} since the position information provide prior knowledge for controlling context interaction.
\textbf{Average, Voting}, and \textbf{Last:} \textbf{Average} takes the average of the local satisfaction distributions of customer utterances for classification. \textbf{Voting} directly maps the majority local satisfaction distributions of customer into satisfaction prediction. \textbf{Last} takes the last customer's satisfaction distribution as classification result. Average, Voting and Last are sub-optimal choices and perform worse than the Full Model. This is because the local satisfaction distributions contribute unequally to the overall satisfaction polarity. Also, the majority satisfaction polarity does not directly correlate with the overall satisfaction.

\subsection{Case Study}
Figure \ref{case} illustrates our prediction results with an example dialogue, which is translated from Chinese text. In this case, three utterances (A$_4$, C$_5$ and C$_6$) are labeled as \textit{transferable}, and two of them (C$_5$ and C$_6$) are labeled as ``negative emotion''. Among them, A$_4$ is an unsatisfactory response, which arouses negative emotions of the customer. DAMI only predicts C$_5$ and C$_6$ as \textit{transferable} utterances. However our model successfully detects all the \textit{transferable} utterances. By mapping local satisfaction distribution of utterances to sentiment of utterances, our model is able to predict reasonable sentiment polarities for customer utterances (detailed analysis is in Subsection \ref{senti_analyze}). Considering the context, the customer describes his/her skin problem at C$_3$ and asks for a recommendation. However, the chatbot does not give any recommendations and returns an irrelevant answer at A$_4$. We provide the attention distributions of the utterances on the right side of the example dialog. ${\alpha}_{5}^{s}$ and ${\alpha}_{6}^{s}$ are the \textit{SSA to MHCH} attention distributions of C$_5$ and C$_6$; ${\alpha}_{5}^{m}$ and ${\alpha}_{6}^{m}$ are the \textit{MHCH to SSA} attention distributions of C$_5$ and C$_6$. We can observe that attention distributions are concentrated on A$_4$ rather than other utterances. It is because A$_4$ is the main cause of negative emotion and dissatisfaction. This again demonstrates that our model can capture the mutual influence between local satisfaction and handoff, which is useful for prediction. In terms of final satisfaction rating, although CAMIL correctly predicts the sentiments of customer utterances, it gives a wrong prediction of satisfaction rating. Our model correctly predicts the satisfaction rating as \textit{Unsatisfied} by considering the negative emotions and its cause of the unsatisfied response.

\subsection{Results on Sentiment Classification}
\label{senti_analyze}
\citet{song2019using} utilize multiple instance learning to predict the satisfaction rating and the sentiment of customer utterances with the supervision of the dialogue's satisfaction labels only during the training process. Similarly, our satisfaction prediction is based on the estimation of local satisfaction distributions while the utterance sentiment or satisfaction labels are unobserved. To compare and analyze the performance of utterance-level sentiment classification, we map these distributions into sentiments of utterances as the sentiment prediction results according to the distribution polarities, i.e., unsatisfied $\rightarrow $ negative (NG), met $\rightarrow $ neutral (NE), well-satisfied $\rightarrow $ positive (PO).

\renewcommand\tabcolsep{2pt}
\begin{table}[t]
  \centering
  \resizebox{0.48\textwidth}{!}{
    \begin{tabular}{l|ccccc|ccccc}
    \toprule
    \multirow{2}[4]{*}{\textbf{Models}} & \multicolumn{5}{c|}{\textbf{Clothes}} & \multicolumn{5}{c}{\textbf{Makeup}} \\
\cmidrule{2-11}          & PO F1 & NE F1 & NG F1 & Mac. F1 & Acc.  & PO F1 & NE F1 & NG F1 & Mac. F1 & Acc. \\
    \midrule
    \midrule
    MILNET & 44.1  & 81.4  & 40.4  & 55.3  & 71.3  & 44.7  & 38.7  & 41.6  & 41.7  & 41.0 \\
    CAMIL & 48.4  & 89.3  & 55.5  & 64.4  & 82.4  & \textbf{54.4}  & \textbf{72.5}  & 51.6  & \textbf{59.5}  & \textbf{64.7} \\
    RSSN  & \textbf{63.5}  & \textbf{90.1}  & \textbf{58.4}  & \textbf{70.7}  & \textbf{83.8}  & 51.3  & 67.8  & \textbf{54.6}  & 57.9  & 61.6 \\
    \bottomrule
    \end{tabular}%
    }
    \caption{Results of sentiment classification by different models on Clothes and Makeup test datasets.}
  \label{tab:senti}%
\end{table}%

In Table \ref{tab:senti}, we compare the sentiment prediction results of MILNET, CAMIL, and our model. On Clothes dataset, our RSSN performs better than other baselines, while it performs worse than CAMIL on Makeup dataset. It is worth noting that our model achieves the best performance on both Clothes and Makeup datasets in terms of NG F1 metric. It indicates that MHCH task is sensitive to negative emotion and contributes more to negative emotion recognition than separate SSA models. From Table \ref{tab:compare}, we can also see that our model performs better than separate SSA models in terms of US F1, which is consistent with the findings of sentiment classification.

\section{Conclusions and Future works}
In this paper, we propose an innovative multi-task framework for service satisfaction analysis and machine-human chatting handoff, which deliberately establishes the mutual interrelation for each other. Specifically, we propose a Role-Selected Sharing Network for joint handoff prediction and satisfaction estimation, utilizing role and positional information to control knowledge transfer for both tasks. Extensive experiments and analyses reveal that explicitly modeling the interrelation between the two tasks can boost the performance mutually.

However, our model has not been calibrated to account for user preferences and biases, which we plan to address in future work. Moreover, we will further explore how to adjust the handoff priority with the assistance of personalized information.

\section*{Acknowledgments}
We thank the anonymous reviewers for their valuable comments and suggestions. This work is supported by the National Natural Science Foundation of China (61876003, 62106039), the National Key R\&D Program of China (2020YFC0832505),  the Fundamental Research Funds for the Central Universities, and Alibaba Group through Alibaba Research Intern Program and Alibaba Research Fellowship Program.

\bibliography{ref}

\begin{thebibliography}{42}
\expandafter\ifx\csname natexlab\endcsname\relax\def\natexlab#1{#1}\fi

\bibitem[{Angelidis and Lapata(2018)}]{angelidis2018multiple}
Stefanos Angelidis and Mirella Lapata. 2018.
\newblock Multiple instance learning networks for fine-grained sentiment
  analysis.
\newblock \emph{TACL}, 6:17--31.

\bibitem[{Ba et~al.(2016)Ba, Kiros, and Hinton}]{ba2016layer}
Jimmy~Lei Ba, Jamie~Ryan Kiros, and Geoffrey~E Hinton. 2016.
\newblock Layer normalization.
\newblock \emph{arXiv preprint arXiv:1607.06450}.

\bibitem[{Bodigutla et~al.(2019{\natexlab{a}})Bodigutla, Polymenakos, and
  Matsoukas}]{bodigutla2019multi}
Praveen~Kumar Bodigutla, Lazaros Polymenakos, and Spyros Matsoukas.
  2019{\natexlab{a}}.
\newblock Multi-domain conversation quality evaluation via user satisfaction
  estimation.
\newblock \emph{arXiv preprint arXiv:1911.08567}.

\bibitem[{Bodigutla et~al.(2020)Bodigutla, Tiwari, Matsoukas, Valls-Vargas, and
  Polymenakos}]{bodigutla2020joint}
Praveen~Kumar Bodigutla, Aditya Tiwari, Spyros Matsoukas, Josep Valls-Vargas,
  and Lazaros Polymenakos. 2020.
\newblock Joint turn and dialogue level user satisfaction estimation on
  mulit-domain conversations.
\newblock In \emph{Proc. of EMNLP: Findings}, pages 3897--3909.

\bibitem[{Bodigutla et~al.(2019{\natexlab{b}})Bodigutla, Wang, Ridgeway, Levy,
  Joshi, Geramifard, and Matsoukas}]{bodigutla2019domain}
Praveen~Kumar Bodigutla, Longshaokan Wang, Kate Ridgeway, Joshua Levy, Swanand
  Joshi, Alborz Geramifard, and Spyros Matsoukas. 2019{\natexlab{b}}.
\newblock Domain-independent turn-level dialogue quality evaluation via user
  satisfaction estimation.
\newblock \emph{arXiv preprint arXiv:1908.07064}.

\bibitem[{Brandtzaeg and F{\o}lstad(2018)}]{brandtzaeg2018chatbots}
Petter~Bae Brandtzaeg and Asbj{\o}rn F{\o}lstad. 2018.
\newblock Chatbots: {C}hanging user needs and motivations.
\newblock \emph{Interactions}, 25(5):38--43.

\bibitem[{Chaves and Gerosa(2020)}]{chaves2020should}
Ana~Paula Chaves and Marco~Aurelio Gerosa. 2020.
\newblock How should my chatbot interact? {A} survey on social characteristics
  in human--chatbot interaction design.
\newblock \emph{IJHCI}, pages 1--30.

\bibitem[{Chung and Park(2019)}]{chung2019chatbot}
Kyungyong Chung and Roy~C Park. 2019.
\newblock Chatbot-based heathcare service with a knowledge base for cloud
  computing.
\newblock \emph{Cluster Computing}, 22(1):1925--1937.

\bibitem[{Dai et~al.(2020)Dai, Fu, Zhu, Cui, Qi et~al.}]{dai2020local}
Zhigang Dai, Jinhua Fu, Qile Zhu, Hengbin Cui, Yuan Qi, et~al. 2020.
\newblock Local contextual attention with hierarchical structure for dialogue
  act recognition.
\newblock \emph{arXiv preprint arXiv:2003.06044}.

\bibitem[{Devlin et~al.(2019)Devlin, Chang, Lee, and
  Toutanova}]{devlin2019bert}
Jacob Devlin, Ming-Wei Chang, Kenton Lee, and Kristina Toutanova. 2019.
\newblock {BERT}: Pre-training of deep bidirectional transformers for language
  understanding.
\newblock In \emph{Proc. of NAACL}, pages 4171--4186.

\bibitem[{El~Asri et~al.(2014)El~Asri, Khouzaimi, Laroche, and
  Pietquin}]{el2014ordinal}
Layla El~Asri, Hatim Khouzaimi, Romain Laroche, and Olivier Pietquin. 2014.
\newblock Ordinal regression for interaction quality prediction.
\newblock In \emph{Proc. of ICASSP}, pages 3221--3225.

\bibitem[{Glorot and Bengio(2010)}]{glorot2010understanding}
Xavier Glorot and Yoshua Bengio. 2010.
\newblock Understanding the difficulty of training deep feedforward neural
  networks.
\newblock In \emph{Proc of AISTATS}, pages 249--256.

\bibitem[{Higashinaka et~al.(2015{\natexlab{a}})Higashinaka, Funakoshi, Araki,
  Tsukahara, Kobayashi, and Mizukami}]{higashinaka2015towards}
Ryuichiro Higashinaka, Kotaro Funakoshi, Masahiro Araki, Hiroshi Tsukahara,
  Yuka Kobayashi, and Masahiro Mizukami. 2015{\natexlab{a}}.
\newblock Towards taxonomy of errors in chat-oriented dialogue systems.
\newblock In \emph{Proc. of SIGDIAL}, pages 87--95.

\bibitem[{Higashinaka et~al.(2015{\natexlab{b}})Higashinaka, Mizukami,
  Funakoshi, Araki, Tsukahara, and Kobayashi}]{higashinaka2015fatal}
Ryuichiro Higashinaka, Masahiro Mizukami, Kotaro Funakoshi, Masahiro Araki,
  Hiroshi Tsukahara, and Yuka Kobayashi. 2015{\natexlab{b}}.
\newblock Fatal or not? {F}inding errors that lead to dialogue breakdowns in
  chat-oriented dialogue systems.
\newblock In \emph{Proc. of EMNLP}, pages 2243--2248.

\bibitem[{Hochreiter and Schmidhuber(1997)}]{hochreiter1997long}
Sepp Hochreiter and J{\"u}rgen Schmidhuber. 1997.
\newblock Long short-term memory.
\newblock \emph{Neural computation}, 9(8):1735--1780.

\bibitem[{Huang et~al.(2018)Huang, Chang, and Bigham}]{huang2018evorus}
Ting{-}Hao~Kenneth Huang, Joseph~Chee Chang, and Jeffrey~P. Bigham. 2018.
\newblock Evorus: {A} crowd-powered conversational assistant built to automate
  itself over time.
\newblock In \emph{Proc. of CHI}, pages 1--13.

\bibitem[{Jain et~al.(2018)Jain, Kumar, Kota, and Patel}]{jain2018evaluating}
Mohit Jain, Pratyush Kumar, Ramachandra Kota, and Shwetak~N Patel. 2018.
\newblock Evaluating and informing the design of chatbots.
\newblock In \emph{Proc. of DIS}, pages 895--906.

\bibitem[{Kingma and Ba(2015)}]{kingma2014adam}
Diederik~P. Kingma and Jimmy Ba. 2015.
\newblock Adam: {A} method for stochastic optimization.
\newblock In \emph{Proc. of ICLR (Poster)}.

\bibitem[{Kumar et~al.(2018)Kumar, Agarwal, Dasgupta, and
  Joshi}]{kumar2017dialogue}
Harshit Kumar, Arvind Agarwal, Riddhiman Dasgupta, and Sachindra Joshi. 2018.
\newblock Dialogue act sequence labeling using hierarchical encoder with {CRF}.
\newblock In \emph{Proc. of AAAI}, pages 3440--3447.

\bibitem[{Liu et~al.(2021)Liu, Gao, Kang, Jiang, He, Sun, Liu, and
  Lu}]{liu2021time}
Jiawei Liu, Zhe Gao, Yangyang Kang, Zhuoren Jiang, Guoxiu He, Changlong Sun,
  Xiaozhong Liu, and Wei Lu. 2021.
\newblock Time to transfer: {P}redicting and evaluating machine-human chatting
  handoff.
\newblock In \emph{Proc. of AAAI}, pages 5841--5849.

\bibitem[{Ma et~al.(2018)Ma, Gao, and Wong}]{ma2018detect}
Jing Ma, Wei Gao, and Kam-Fai Wong. 2018.
\newblock Detect rumor and stance jointly by neural multi-task learning.
\newblock In \emph{Proc. of WWW}, pages 585--593.

\bibitem[{Majumder et~al.(2019)Majumder, Poria, Hazarika, Mihalcea, Gelbukh,
  and Cambria}]{majumder2019dialoguernn}
Navonil Majumder, Soujanya Poria, Devamanyu Hazarika, Rada Mihalcea,
  Alexander~F. Gelbukh, and Erik Cambria. 2019.
\newblock Dialoguernn: An attentive {RNN} for emotion detection in
  conversations.
\newblock In \emph{Proc. of AAAI}, pages 6818--6825.

\bibitem[{Mikolov et~al.(2013)Mikolov, Chen, Corrado, and
  Dean}]{mikolov2013efficient}
Tomas Mikolov, Kai Chen, Greg Corrado, and Jeffrey Dean. 2013.
\newblock Efficient estimation of word representations in vector space.
\newblock \emph{arXiv preprint arXiv:1301.3781}.

\bibitem[{Qin et~al.(2020)Qin, Che, Li, Ni, and Liu}]{qin2020dcr}
Libo Qin, Wanxiang Che, Yangming Li, Minheng Ni, and Ting Liu. 2020.
\newblock {DCR-N}et: A deep co-interactive relation network for joint dialog
  act recognition and sentiment classification.
\newblock In \emph{Proc. of AAAI}, pages 8665--8672.

\bibitem[{Qiu et~al.(2017)Qiu, Li, Wang, Gao, Chen, Zhao, Chen, Huang, and
  Chu}]{qiu2017alime}
Minghui Qiu, Feng-Lin Li, Siyu Wang, Xing Gao, Yan Chen, Weipeng Zhao, Haiqing
  Chen, Jun Huang, and Wei Chu. 2017.
\newblock {A}li{M}e chat: A sequence to sequence and rerank based chatbot
  engine.
\newblock In \emph{Proc. of ACL}, pages 498--503.

\bibitem[{Radziwill and Benton(2017)}]{radziwill2017evaluating}
Nicole Radziwill and Morgan Benton. 2017.
\newblock Evaluating quality of chatbots and intelligent conversational agents.
\newblock \emph{Software Quality Professional}, 19(3):25.

\bibitem[{Raheja and Tetreault(2019)}]{raheja2019dialogue}
Vipul Raheja and Joel Tetreault. 2019.
\newblock {D}ialogue {A}ct {C}lassification with {C}ontext-{A}ware
  {S}elf-{A}ttention.
\newblock In \emph{Proc. of NAACL}, pages 3727--3733.

\bibitem[{Rajendran et~al.(2019)Rajendran, Ganhotra, and
  Polymenakos}]{rajendran2019learning}
Janarthanan Rajendran, Jatin Ganhotra, and Lazaros~C. Polymenakos. 2019.
\newblock Learning end-to-end goal-oriented dialog with maximal user task
  success and minimal human agent use.
\newblock \emph{TACL}, 7:375--386.

\bibitem[{Ram et~al.(2018)Ram, Prasad, Khatri, Venkatesh, Gabriel, Liu, Nunn,
  Hedayatnia, Cheng, Nagar et~al.}]{ram2018conversational}
Ashwin Ram, Rohit Prasad, Chandra Khatri, Anu Venkatesh, Raefer Gabriel, Qing
  Liu, Jeff Nunn, Behnam Hedayatnia, Ming Cheng, Ashish Nagar, et~al. 2018.
\newblock Conversational {AI}: The science behind the alexa prize.
\newblock \emph{arXiv preprint arXiv:1801.03604}.

\bibitem[{Schmitt et~al.(2012)Schmitt, Ultes, and
  Minker}]{schmitt2012parameterized}
Alexander Schmitt, Stefan Ultes, and Wolfgang Minker. 2012.
\newblock A parameterized and annotated spoken dialog corpus of the {CMU}
  let{'}s go bus information system.
\newblock In \emph{Proc. of LREC}, pages 3369--3373.

\bibitem[{Shen et~al.(2018)Shen, Sun, Wang, Kang, Li, Liu, Si, Zhang, and
  Zhou}]{shen2018sentiment}
Chenlin Shen, Changlong Sun, Jingjing Wang, Yangyang Kang, Shoushan Li,
  Xiaozhong Liu, Luo Si, Min Zhang, and Guodong Zhou. 2018.
\newblock Sentiment classification towards question-answering with hierarchical
  matching network.
\newblock In \emph{Proc. of EMNLP}, pages 3654--3663.

\bibitem[{Snow et~al.(2008)Snow, O{'}Connor, Jurafsky, and Ng}]{snow2008cheap}
Rion Snow, Brendan O{'}Connor, Daniel Jurafsky, and Andrew Ng. 2008.
\newblock Cheap and fast {--} but is it good? evaluating non-expert annotations
  for natural language tasks.
\newblock In \emph{Proc. of EMNLP}, pages 254--263.

\bibitem[{Song et~al.(2019)Song, Bing, Gao, Lin, Zhao, Wang, Sun, Liu, and
  Zhang}]{song2019using}
Kaisong Song, Lidong Bing, Wei Gao, Jun Lin, Lujun Zhao, Jiancheng Wang,
  Changlong Sun, Xiaozhong Liu, and Qiong Zhang. 2019.
\newblock Using customer service dialogues for satisfaction analysis with
  context-assisted multiple instance learning.
\newblock In \emph{Proc. of EMNLP}, pages 198--207.

\bibitem[{Srivastava et~al.(2014)Srivastava, Hinton, Krizhevsky, Sutskever, and
  Salakhutdinov}]{srivastava2014dropout}
Nitish Srivastava, Geoffrey Hinton, Alex Krizhevsky, Ilya Sutskever, and Ruslan
  Salakhutdinov. 2014.
\newblock Dropout: a simple way to prevent neural networks from overfitting.
\newblock \emph{JMLR}, 15(1):1929--1958.

\bibitem[{Ultes(2019)}]{ultes2019improving}
Stefan Ultes. 2019.
\newblock Improving interaction quality estimation with {B}i{LSTM}s and the
  impact on dialogue policy learning.
\newblock In \emph{Proc. of SIGDIAL}, pages 11--20.

\bibitem[{Ultes et~al.(2014)Ultes, ElChab, and Minker}]{ultes2014application}
Stefan Ultes, Robert ElChab, and Wolfgang Minker. 2014.
\newblock Application and evaluation of a conditioned hidden markov model for
  estimating interaction quality of spoken dialogue systems.
\newblock \emph{Natural Interaction with Robots, Knowbots and Smartphones},
  pages 303--312.

\bibitem[{Vaswani et~al.(2017)Vaswani, Shazeer, Parmar, Uszkoreit, Jones,
  Gomez, Kaiser, and Polosukhin}]{vaswani2017attention}
Ashish Vaswani, Noam Shazeer, Niki Parmar, Jakob Uszkoreit, Llion Jones,
  Aidan~N. Gomez, Lukasz Kaiser, and Illia Polosukhin. 2017.
\newblock Attention is all you need.
\newblock In \emph{Proc. of NIPS}, pages 5998--6008.

\bibitem[{Wang et~al.(2020{\natexlab{a}})Wang, Zhang, Liu, Sun, and
  Zhang}]{wang2020masking}
Tianyi Wang, Yating Zhang, Xiaozhong Liu, Changlong Sun, and Qiong Zhang.
  2020{\natexlab{a}}.
\newblock Masking orchestration: Multi-task pretraining for multi-role dialogue
  representation learning.
\newblock In \emph{Proc. of AAAI}, pages 9217--9224.

\bibitem[{Wang et~al.(2020{\natexlab{b}})Wang, Zhang, Ma, Wang, and
  Xiao}]{wang2020contextualized}
Yan Wang, Jiayu Zhang, Jun Ma, Shaojun Wang, and Jing Xiao. 2020{\natexlab{b}}.
\newblock Contextualized emotion recognition in conversation as sequence
  tagging.
\newblock In \emph{Proc. of SIGDIAL}, pages 186--195.

\bibitem[{Xiong et~al.(2016)Xiong, Merity, and Socher}]{xiong2016dynamic}
Caiming Xiong, Stephen Merity, and Richard Socher. 2016.
\newblock Dynamic memory networks for visual and textual question answering.
\newblock In \emph{Proc. of ICML}, pages 2397--2406.

\bibitem[{Yang et~al.(2016)Yang, Yang, Dyer, He, Smola, and
  Hovy}]{yang2016hierarchical}
Zichao Yang, Diyi Yang, Chris Dyer, Xiaodong He, Alex Smola, and Eduard Hovy.
  2016.
\newblock Hierarchical attention networks for document classification.
\newblock In \emph{Proc. of NAACL}, pages 1480--1489.

\bibitem[{Zhou et~al.(2020)Zhou, Gao, Li, and Shum}]{zhou2020design}
Li~Zhou, Jianfeng Gao, Di~Li, and Heung-Yeung Shum. 2020.
\newblock The design and implementation of {X}iao{I}ce, an empathetic social
  chatbot.
\newblock \emph{Computational Linguistics}, 46(1):53--93.

\end{thebibliography}
\bibliographystyle{acl_natbib}

\end{document}